\PassOptionsToPackage{unicode}{hyperref}
\PassOptionsToPackage{hyphens}{url}
\documentclass[
  11pt,
]{article}
\usepackage{amsmath,amssymb}
\usepackage{iftex}
\ifPDFTeX
  \usepackage[T1]{fontenc}
  \usepackage[utf8]{inputenc}
  \usepackage{textcomp} 
\else 
  \usepackage{unicode-math} 
  \defaultfontfeatures{Scale=MatchLowercase}
  \defaultfontfeatures[\rmfamily]{Ligatures=TeX,Scale=1}
\fi
\usepackage{lmodern}
\ifPDFTeX\else
\fi
\IfFileExists{upquote.sty}{\usepackage{upquote}}{}
\IfFileExists{microtype.sty}{
  \usepackage[]{microtype}
  \UseMicrotypeSet[protrusion]{basicmath} 
}{}
\makeatletter
\@ifundefined{KOMAClassName}{
  \IfFileExists{parskip.sty}{%
    \usepackage{parskip}
  }{
    \setlength{\parindent}{0pt}
    \setlength{\parskip}{6pt plus 2pt minus 1pt}}
}{
  \KOMAoptions{parskip=half}}
\makeatother
\usepackage{xcolor}
\usepackage[margin=0.85in]{geometry}
\providecommand{\tightlist}{%
  \setlength{\itemsep}{0pt}\setlength{\parskip}{0pt}}
\usepackage{xurl}
\usepackage{fancyhdr}
\usepackage{microtype}
\usepackage{needspace}
\usepackage{newunicodechar}
\newunicodechar{∈}{\ensuremath{\in}}
\newunicodechar{π}{\ensuremath{\pi}}
\newunicodechar{δ}{\ensuremath{\delta}}
\newunicodechar{≤}{\ensuremath{\leq}}
\newunicodechar{≥}{\ensuremath{\geq}}
\newunicodechar{ε}{\ensuremath{\epsilon}}
\newunicodechar{φ}{\ensuremath{\varphi}}
\newunicodechar{κ}{\ensuremath{\kappa}}
\newunicodechar{Γ}{\ensuremath{\Gamma}}
\newunicodechar{Σ}{\ensuremath{\Sigma}}
\newunicodechar{Δ}{\ensuremath{\Delta}}
\newunicodechar{√}{\ensuremath{\surd}}
\newunicodechar{ⱼ}{\ensuremath{{}_j}}
\newunicodechar{ₖ}{\ensuremath{{}_k}}
\newunicodechar{ₘ}{\ensuremath{{}_m}}
\newunicodechar{ₙ}{\ensuremath{{}_n}}
\newunicodechar{ₜ}{\ensuremath{{}_t}}
\newunicodechar{₀}{\ensuremath{{}_0}}
\newunicodechar{ᵣ}{\ensuremath{{}_r}}
\newunicodechar{ᵈ}{\ensuremath{{}^{d}}}
\newunicodechar{ᵀ}{\ensuremath{{}^{T}}}
\newunicodechar{⁻}{\ensuremath{{}^{-}}}
\newunicodechar{ℝ}{\ensuremath{\mathbb{R}}}
\newunicodechar{ℓ}{\ensuremath{\ell}}
\ifLuaTeX
  \usepackage{selnolig}  
\fi
\IfFileExists{bookmark.sty}{\usepackage{bookmark}}{\usepackage{hyperref}}
\IfFileExists{xurl.sty}{\usepackage{xurl}}{} 
\hypersetup{
  pdftitle={Phase Structure in Rotary Attention},
  pdfauthor={Abraham Chachamovits},
  hidelinks,
  pdfcreator={LaTeX via pandoc}}

\title{Phase Structure in Rotary Attention}
\usepackage{etoolbox}
\makeatletter
\providecommand{\subtitle}[1]{
  \apptocmd{\@title}{\par {\large #1 \par}}{}{}
}
\makeatother
\subtitle{A Spectral Framework for Semantic Continuity and
Execution-Boundary Governance}
\author{Abraham Chachamovits}
\date{ENTRUST AI ·
\href{mailto:contact@entrustai.co}{\nolinkurl{contact@entrustai.co}} ·
September 2026}

\begin{document}
\maketitle

\hypertarget{abstract}{%
\section{Abstract}\label{abstract}}

Transformer language models are usually analyzed through vector
geometry, yet ordered context and rotary position encoding introduce
explicit phase structure into query--key interactions. This paper
develops a spectral framework for examining rotary phase alignment and
hidden-state continuity without treating language models as literal
physical wave systems. It identifies ordered hidden-state sequences,
rather than arbitrary vocabulary indices, as domains for spectral
decomposition. It expresses the Rotary Position Embedding (RoPE)
attention score as a sum of magnitude-weighted cosine terms and derives
a local score-stability bound: uniformly bounded phase displacement
limits loss relative to full alignment when pair magnitudes are fixed.
To extend analysis beyond native RoPE coordinates, it defines complex
modal coordinates over fixed orthonormal direction pairs and a weighted
coherence functional with explicit domain conditions. The predictive
usefulness of that functional requires empirical evidence beyond its
mathematical definition. The paper separately specifies an idealized
distinction between representational continuity and execution-boundary
admissibility: internal coherence cannot authorize a consequential
transition.

Keywords: transformers; rotary position embedding; RoPE; phase geometry;
spectral analysis; semantic continuity; representation drift; AI
governance; execution boundaries

\hypertarget{introduction}{%
\section{1. Introduction}\label{introduction}}

Modern language models are ordinarily described through geometry. Tokens
are embedded as vectors, transformed through successive layers, compared
by inner products, and converted into probability distributions over
possible continuations. That account identifies the spaces in which
representations live and the operators through which they are
transformed.

Transformer computation also contains ordered trajectories and
positional rotations. Rotary Position Embedding implements position as
coordinated rotation, with relative displacement appearing directly in
the query--key interaction {[}2{]}. This permits a phase-coordinate
description of compatibility. Whether a related modal description of
hidden states provides useful predictors of semantic continuity is a
separate question.

\hypertarget{related-work-and-contribution-boundary}{%
\subsection{1.1 Related work and contribution
boundary}\label{related-work-and-contribution-boundary}}

The framework lies at the intersection of Transformer attention, rotary
position encoding, representation analysis, and mechanistic
interpretability. Transformer attention combines content-dependent
query--key compatibility with normalized allocation {[}1{]}. RoPE makes
relative position algebraically available through blockwise rotations
{[}2{]}. Representation-analysis methods compare learned spaces rather
than assuming that arbitrary coordinates have intrinsic meaning {[}4{]}.
Mechanistic interpretability investigates computations implemented by
model components, including induction-head behavior {[}5{]}.

Barbero et al.~{[}6{]} identify robust positional attention patterns
associated with high RoPE frequencies in Gemma and hypothesize a
semantic role for the preferred lower frequencies. Gu et al.~{[}7{]}
analyze how additive and multiplicative positional encodings combine
positional and semantic information in attention logits. Liu {[}8{]}
develops a phase-modulation account with theoretical bounds relating the
RoPE base to context length, depth, and numerical precision. These
studies address specific properties of positional encoding; the present
proposal separately asks whether paired hidden-state coordinates provide
useful continuity predictors. This paper does not claim the first phase
interpretation of RoPE, the first spectral analysis of positional
encoding, a new encoding scheme, or an empirically discovered circuit.
The rotary decomposition below restates the standard algebra in complex
coordinates; the stability lemma is a direct consequence of an
elementary cosine inequality.

The methodological proposal connects that decomposition and bound with a
reproducible paired-coordinate construction for hidden-state continuity
analysis. It also specifies an architectural distinction between
continuity and authorization. The contribution is a methodological
synthesis whose predictive value depends on comparative evidence.

The terms \emph{spectral}, \emph{phase}, \emph{mode}, and
\emph{coherence} refer to mathematical representations of ordered
computational states. They do not imply propagating energy, quantum
behavior, or literal oscillation in physical space.

\hypertarget{spectral-structure-belongs-to-ordered-context}{%
\section{2. Spectral Structure Belongs to Ordered
Context}\label{spectral-structure-belongs-to-ordered-context}}

A vocabulary index is an arbitrary identifier. Its numerical order
supplies no intrinsic neighborhood, distance, or frequency. A transform
over that arbitrary order would not acquire semantic meaning merely by
being a Fourier transform. An appropriate ordered domain is the sequence
processed by the model.

Let a transformer layer ℓ produce hidden representations

\[
h_0^{(\ell)},h_1^{(\ell)},\ldots,h_{L-1}^{(\ell)}\in\mathbb R^d. \tag{1}
\]

Collect them into

\[
H^{(\ell)}=\begin{pmatrix}(h_0^{(\ell)})^\top\\(h_1^{(\ell)})^\top\\\vdots\\(h_{L-1}^{(\ell)})^\top\end{pmatrix}\in\mathbb R^{L\times d}. \tag{2}
\]

The rows follow token position within the active context. For a
representational direction u ∈ ℝᵈ, define

\[
y_m^{(\ell,u)}=\langle h_m^{(\ell)},u\rangle,\qquad m=0,\ldots,L-1. \tag{3}
\]

This ordered sequence admits the discrete Fourier representation

\[
\widehat y_k^{(\ell,u)}=\sum_{m=0}^{L-1}y_m^{(\ell,u)}e^{-2\pi i km/L},\qquad
y_m^{(\ell,u)}=\frac1L\sum_{k=0}^{L-1}\widehat y_k^{(\ell,u)}e^{2\pi i km/L}. \tag{4}
\]

A vector-valued transform may likewise be defined:

\[
\widehat h_k^{(\ell)}=\sum_{m=0}^{L-1}h_m^{(\ell)}e^{-2\pi i km/L},\qquad
h_m^{(\ell)}=\frac1L\sum_{k=0}^{L-1}\widehat h_k^{(\ell)}e^{2\pi i km/L}. \tag{5}
\]

This decomposition does not claim that the model explicitly computes a
Fourier transform. It provides coordinates for positional variation in
an ordered representation sequence. Spectral analysis belongs to the
broader mathematical study of representations through frequency- or
scale-localized components {[}3{]}.

Slowly varying modes may capture broad contextual structure, while
higher-frequency modes may describe local alternation, abrupt
transitions, repetition, or short-range positional variation. Those
semantic interpretations are hypotheses, not consequences of the
existence of the transform. The object analyzed is the ordered
representational sequence, not an isolated token's arbitrary vocabulary
identifier.

\hypertarget{rotary-position-embedding-as-phase-geometry}{%
\section{3. Rotary Position Embedding as Phase
Geometry}\label{rotary-position-embedding-as-phase-geometry}}

The original Transformer replaced recurrence and convolution with
attention-based sequence processing {[}1{]}. RoPE introduces a
particular positional structure through rotations in paired coordinate
planes {[}2{]}. For an even-dimensional query or key vector, define

\[
R_{\theta_j,m}=\begin{pmatrix}\cos(m\theta_j)&-\sin(m\theta_j)\\\sin(m\theta_j)&\cos(m\theta_j)\end{pmatrix}. \tag{6}
\]

The full rotary operator is

\[
R^d_{\Theta,m}=\operatorname{diag}(R_{\theta_1,m},\ldots,R_{\theta_{d/2},m}). \tag{7}
\]

Let qⱼ and kⱼ denote the j-th real coordinate pairs. Orthogonality gives

\[
R_{\theta_j,m}^{\top}R_{\theta_j,n}=R_{\theta_j,n-m}, \tag{8}
\]

and hence

\[
\langle R_{\theta_j,m}q_j,R_{\theta_j,n}k_j\rangle
=\langle q_j,R_{\theta_j,n-m}k_j\rangle. \tag{9}
\]

Identify each pair with a complex scalar:

\[
q_j=q_{j,1}+i q_{j,2},\qquad k_j=k_{j,1}+i k_{j,2}. \tag{10}
\]

The rotations become

\[
q_j^{(m)}=q_j e^{im\theta_j},\qquad k_j^{(n)}=k_j e^{in\theta_j}. \tag{11}
\]

Their real inner product is

\[
\langle R_{\theta_j,m}q_j,R_{\theta_j,n}k_j\rangle
=\operatorname{Re}\!\left(q_j\overline{k_j}e^{i(m-n)\theta_j}\right). \tag{12}
\]

For nonzero coordinates, write \(q_j=|q_j|e^{i\alpha_j}\) and
\(k_j=|k_j|e^{i\beta_j}\). Then

\[
\langle R_{\theta_j,m}q_j,R_{\theta_j,n}k_j\rangle
=|q_j||k_j|\cos\!\left(\alpha_j-\beta_j+(m-n)\theta_j\right). \tag{13}
\]

A zero coordinate contributes zero through Eq. (12), without requiring a
phase assignment. Summing over pairs yields the unscaled score:

\[
\langle R^d_{\Theta,m}q,R^d_{\Theta,n}k\rangle
=\sum_{j=1}^{d/2}|q_j||k_j|\cos\!\left(\alpha_j-\beta_j+(m-n)\theta_j\right). \tag{14}
\]

Phase here is the angular coordinate of a two-dimensional pair. Each
nonzero pair contributes according to its content-dependent query and
key phases and its relative-position phase. These equations assume that
the full stated query--key dimension is rotary; nonrotary coordinates,
if present in an architecture, contribute their ordinary inner-product
terms separately.

\hypertarget{modal-reinforcement-and-attenuation}{%
\section{4. Modal Reinforcement and
Attenuation}\label{modal-reinforcement-and-attenuation}}

For a fully rotary attention head h with even query--key dimension dₖ,
the scaled score is

\[
S_{mn}^{(h)}=\frac{\langle R^{d_k}_{\Theta,m}q_m^{(h)},R^{d_k}_{\Theta,n}k_n^{(h)}\rangle}{\sqrt{d_k}}. \tag{15}
\]

Equivalently,

\[
\begin{aligned}
S_{mn}^{(h)}&=\frac1{\sqrt{d_k}}\sum_{j=1}^{d_k/2}\rho_{mnj}^{(h)}\cos\delta_{mnj}^{(h)},\\
\rho_{mnj}^{(h)}&=|q_{m,j}^{(h)}||k_{n,j}^{(h)}|,\\
\delta_{mnj}^{(h)}&=\alpha_{m,j}^{(h)}-\beta_{n,j}^{(h)}+(m-n)\theta_j.
\end{aligned}\tag{16}
\]

A pair near zero phase displacement modulo 2π contributes near its
maximum. Near π/2 its contribution is small; near π it is negative. The
score combines these contributions algebraically, without implying
physical wave propagation.

For causal self-attention,

\[
M_{mn}=\begin{cases}0,&n\le m,\\-\infty,&n>m.\end{cases}\tag{17}
\]

Masked softmax gives

\[
A_{mn}^{(h)}=\frac{\exp(S_{mn}^{(h)}+M_{mn})}{\sum_{r=0}^{L-1}\exp(S_{mr}^{(h)}+M_{mr})}. \tag{18}
\]

Phase structure describes compatibility before competitive allocation.
The normalized weight depends on every permitted competitor in the row.
Attention is therefore not a classical linear time-invariant spectral
filter: its weights depend on content, layer, head, and context.

\hypertarget{a-rope-score-stability-lemma}{%
\section{5. A RoPE Score-Stability
Lemma}\label{a-rope-score-stability-lemma}}

Let

\[
S=\sum_{j=1}^{J}\rho_j\cos\delta_j,\qquad \rho_j\ge0, \tag{19}
\]

and define the fully aligned reference

\[
S_{\max}=\sum_{j=1}^{J}\rho_j. \tag{20}
\]

For a nonzero contributing pair, δⱼ may be taken as the wrapped
displacement from full alignment. Zero-weight terms need no phase
assignment.

\textbf{Lemma 1 (bounded phase displacement bounds score loss).} If
\textbar δⱼ\textbar{} ≤ ε for every contributing pair, with the
magnitudes held fixed, then

\[
S\ge S_{\max}-\frac{\varepsilon^2}{2}\sum_{j=1}^{J}\rho_j
=\left(1-\frac{\varepsilon^2}{2}\right)S_{\max}. \tag{21}
\]

\textbf{Proof.} For every real x,

\[
\cos x\ge1-\frac{x^2}{2}. \tag{22}
\]

Multiplication by a nonnegative weight gives

\[
\rho_j\cos\delta_j\ge\rho_j\left(1-\frac{\delta_j^2}{2}\right). \tag{23}
\]

Since \textbar δⱼ\textbar{} ≤ ε,

\[
\rho_j\cos\delta_j\ge\rho_j\left(1-\frac{\varepsilon^2}{2}\right). \tag{24}
\]

Summing yields

\[
S\ge\left(1-\frac{\varepsilon^2}{2}\right)\sum_j\rho_j
=\left(1-\frac{\varepsilon^2}{2}\right)S_{\max}. \tag{25}
\]

The inequality holds generally under its assumptions and is informative
as a quadratic loss bound near full alignment. The aligned reference
uses the same magnitudes; it need not be attainable by varying position
alone for a given query and key. For the scaled score, both sides are
divided by √dₖ.

The lemma does not bound the final attention probability without
assumptions about competing logits. It also does not, by itself,
establish downstream semantic continuity across heads, layers, or
generated tokens.

\hypertarget{hidden-state-trajectories-and-valid-phase-coordinates}{%
\section{6. Hidden-State Trajectories and Valid Phase
Coordinates}\label{hidden-state-trajectories-and-valid-phase-coordinates}}

A paired-coordinate construction can be applied to hidden-state
trajectories after its basis has been specified. Let hₜ ∈ ℝᵈ denote a
selected representation at decoding step t: for example, a state at the
current position or an explicitly defined pooled-prefix observable.
Choose 2K orthonormal directions, with 2K ≤ d:

\[
\langle u_k,u_r\rangle=\delta_{kr},\quad
\langle v_k,v_r\rangle=\delta_{kr},\quad
\langle u_k,v_r\rangle=0. \tag{26}
\]

Define

\[
z_{k,t}=\langle h_t,u_k\rangle+i\langle h_t,v_k\rangle, \tag{27}
\]

with amplitude and, for nonzero amplitude, phase

\[
\begin{aligned}
A_{k,t}&=\sqrt{\langle h_t,u_k\rangle^2+\langle h_t,v_k\rangle^2},\\
\phi_{k,t}&=\operatorname{atan2}(\langle h_t,v_k\rangle,\langle h_t,u_k\rangle).
\end{aligned}\tag{28}
\]

Phase is defined only when Aₖ,ₜ \textgreater{} 0. Possible constructions
include paired principal or singular directions and task-trained
orthogonal directions. Orthonormal Fourier sine--cosine pairs defined
over a fixed, declared ordering of hidden-state coordinates can also
serve as an analytic control. Such pairs lie in the hidden-state feature
space, not the token-position domain transformed in Section 2; their
frequency labels have no intrinsic positional or semantic meaning. Any
construction derived from operator modes must also satisfy Eq. (26);
arbitrary eigenvectors need not do so. No basis is universally
privileged. Its usefulness depends on reproducibility and demonstrated
relevance to the endpoint being studied.

\hypertarget{basis-dependence-identifiability-and-domain}{%
\subsection{6.1 Basis dependence, identifiability, and
domain}\label{basis-dependence-identifiability-and-domain}}

Absolute modal phase depends on the orientation of a pair. A common
rotation within a fixed pair at every observation shifts both phases
equally and leaves their difference unchanged. The coherence functional
can nevertheless depend on the selected two-dimensional subspaces, their
pairing, and their weights. Basis construction and pairing must
therefore be specified reproducibly before evaluation. Orientation
conventions remain relevant to absolute coordinates and interventions;
they are not needed to cancel a common within-pair rotation in a phase
difference.

For two nonzero modal coordinates,

\[
\Delta\phi_k(t,r)=\operatorname{Arg}\!\left(z_{k,t}\overline{z_{k,r}}\right). \tag{29}
\]

Take fixed weights wₖ ≥ 0 and require D(t,r) = Σₖ wₖAₖ,ₜAₖ,ᵣ
\textgreater{} 0. Define

\[
\Gamma(t,r)=\frac{\sum_{k=1}^{K}w_k A_{k,t}A_{k,r}\cos\Delta\phi_k(t,r)}{\sum_{k=1}^{K}w_k A_{k,t}A_{k,r}}. \tag{30}
\]

When either amplitude is zero, define its numerator contribution as
\(\operatorname{Re}(z_{k,t}\overline{z_{k,r}})=0\), without assigning a
phase to the zero coordinate. With nonnegative weights and positive
denominator,

\[
-1\le\Gamma(t,r)\le1. \tag{31}
\]

This follows by bounding the absolute inner product of each projected
pair by its product of norms and summing with nonnegative weights. Cases
with D(t,r) = 0 require an explicit exclusion or missingness rule fixed
before evaluation.

A value near 1 indicates aggregate alignment in the chosen pairs. A
value near 0 can reflect approximately orthogonal pairs or cancellation
between aligned and opposed contributions; it does not imply that every
pair is weakly aligned. Negative values indicate aggregate opposition.
None of these values automatically establishes semantic fidelity. That
interpretation requires empirical evidence connecting the chosen
construction to the relevant objective, facts, constraints, or task
structure.

\hypertarget{autoregressive-inference-as-an-execution-trajectory}{%
\section{7. Autoregressive Inference as an Execution
Trajectory}\label{autoregressive-inference-as-an-execution-trajectory}}

A transformer is not a conventional recurrent network, but
autoregressive inference produces a sequence of execution states. Let
x≤ₜ = (x₀, \ldots, xₜ) be the generated prefix, and let Eₜ contain the
state required to continue inference. Depending on implementation, this
may include the prefix, cached keys and values, decoding parameters,
active instructions, and external context.

The next-token distribution is

\[
p_\Theta(x_{t+1}\mid E_t). \tag{32}
\]

Under sampling,

\[
x_{t+1}\sim p_\Theta(\,\cdot\mid E_t), \tag{33}
\]

and the state advances according to

\[
E_{t+1}=T_\Theta(E_t,x_{t+1}). \tag{34}
\]

Other decoding rules select tokens differently. The transition is an
execution-state description, not a claim that the transformer has one
recurrent hidden vector. A selected observable is

\[
h_t=O(E_t). \tag{35}
\]

The sequence \(h_0,\ldots,h_T\) forms an analyzable representational
trajectory. It is affected by attention, residual pathways,
normalization, nonlinear activation, token selection, context growth,
and external retrieval or tool results. A spectral account supplies
coordinates for studying this process; it does not turn it into a
conservative oscillator.

\hypertarget{semantic-continuity-and-modal-drift}{%
\section{8. Semantic Continuity and Modal
Drift}\label{semantic-continuity-and-modal-drift}}

A continuation can remain locally fluent while departing from its
initiating objective. Local next-token compatibility does not guarantee
preservation of a global relation. Let h₀ be an anchor associated with a
directive, established facts, or an objective, and let hₜ be the later
observable. A geometric comparison in a selected subspace with projector
P is

\[
\kappa_t=\frac{\langle Ph_t,Ph_0\rangle}{\|Ph_t\|\,\|Ph_0\|}. \tag{36}
\]

This is defined only when both projected norms are nonzero; undefined
cases require a rule fixed before evaluation. The paired construction
gives

\[
H_t=\Gamma(t,0). \tag{37}
\]

Both quantities are geometric functions of the selected representations.
For each pair, Aₖ,ₜAₖ,₀ cos Δφₖ(t,0) is the inner product of the two
projected coordinate vectors. The modal construction normalizes the
weighted sum of these inner products by the sum of paired amplitude
products. This generally differs from the global norm-product
normalization in κₜ. With a single pair and the same projected subspace,
the two coincide. Phase notation alone does not establish additional
information; whether pairing and normalization improve prediction is
empirical.

A trajectory may preserve local compatibility while accumulating
divergence from its anchor. Topic drift might be associated with
displacement in subject-related modes; contradiction with changes among
configurations associated with incompatible commitments; repetition with
persistent concentration; and unsupported elaboration with divergence
from evidence-related constraints despite thematic continuity.

These are empirical hypotheses. They require a basis that distinguishes
the relevant failures from ordinary variation and comparisons against
simpler geometric predictors. The framework's scientific value rests on
comparative prediction, not descriptive elegance.

\hypertarget{coherence-is-not-admissibility}{%
\section{9. Coherence Is Not
Admissibility}\label{coherence-is-not-admissibility}}

Internal coherence and institutional permission are different
properties. An output can be fluent and contextually consistent while
exceeding delegated authority or violating a declared constraint.

Let sₜ denote the governed interaction state and aₜ a candidate output
or action. Specify a governance contract as

\[
C(s_t,a_t)=\bigwedge_{r=1}^{N}p_r(s_t,a_t), \tag{38}
\]

where predicates may represent objective boundaries, scope restrictions,
authority limits, resolved facts, commitments, continuity requirements,
or forbidden transitions. The idealized characteristic decision is

\[
\chi_C(s_t,a_t)=\begin{cases}1,&C(s_t,a_t)=\mathrm{true},\\0,&C(s_t,a_t)=\mathrm{false}.\end{cases}\tag{39}
\]

The executed candidate is

\[
a_t^{\mathrm{exec}}=\begin{cases}a_t,&\chi_C(s_t,a_t)=1,\\\varnothing,&\chi_C(s_t,a_t)=0.\end{cases}\tag{40}
\]

Where correction is supported, let

\[
\widetilde a_t=\mathcal R(s_t,a_t). \tag{41}
\]

The corrected candidate is evaluated again:

\[
a_t^{\mathrm{exec}}=\begin{cases}\widetilde a_t,&\chi_C(s_t,\widetilde a_t)=1,\\\varnothing,&\chi_C(s_t,\widetilde a_t)=0.\end{cases}\tag{42}
\]

Equations (38)--(42) specify an idealized rule conditional on correctly
evaluated predicates and a gate that controls the relevant execution
path. They do not prove evaluator accuracy, institutional legitimacy, or
complete runtime enforcement. An implementation must specify how
unresolved evaluations are handled; under a requirement to establish
admissibility before release, unresolved evaluation cannot count as
acceptance. A model-based semantic assessment is an input to the
decision procedure, not an infallible evaluation of the intended
predicate.

The formulation does not assume a nearest-admissible projection. It
represents evaluation, acceptance, correction where supported, and
blocking where admissibility is not established.

The Continuity Governance Mesh (CGM), developed by ENTRUST AI,
implements an external governance architecture around model-generated
candidates. Its server and interface maintain contract and session
records, evaluate outputs against declared constraints, and provide
correction, reevaluation, holds, and attributed release according to the
configured governance posture. The implementation combines model-based
assessments with deterministic rule checks and integrates The Pilcrow™
for downstream adjudication. Session signing uses HMAC-SHA256 to
authenticate a payload containing contract and ordered-ledger
fingerprints and an explicit seal scope. CGM is included as an
implementation example of the architectural separation between
generation and authorization. This paper does not establish its
end-to-end enforcement effectiveness, and the phase-prediction
experiments do not evaluate that effectiveness.

The conceptual distinction is

\[
\begin{aligned}
\text{semantic coherence}&\not\Rightarrow\text{institutional admissibility},\\
\text{institutional admissibility}&\not\Rightarrow\text{maximal semantic similarity}.
\end{aligned}\tag{43}
\]

A governance contract may permit substantial variation in wording and
reasoning while refusing a particular unauthorized consequence. Its
legitimacy and enforcement require evidence beyond the coherence of the
model's output.

\hypertarget{boundary-stabilization-without-internal-suppression}{%
\section{10. Boundary Stabilization Without Internal
Suppression}\label{boundary-stabilization-without-internal-suppression}}

The architectural objective is to control the transition into accepted
consequence without requiring elimination of internal variation. Let
T(sₜ) be the candidate transitions available from state sₜ, and

\[
\mathcal A(s_t)\subseteq T(s_t) \tag{44}
\]

the subset permitted by the contract. The intended separation is

\[
\text{candidate diversity within }T(s_t),\qquad
\text{executed consequence within }\mathcal A(s_t). \tag{45}
\]

This is a design condition, not a demonstrated guarantee of any
implementation. It does not demand a deterministic latent process. It
requires effective control over the relevant execution boundary, with
the predicate and coverage limitations stated in Section 9.

An external governance system may compare an interaction with an
authorized reference and intervene on departures. This does not
literally phase-lock the model's internal coordinates. Spectral analysis
may describe a trajectory; governance determines which consequences the
institution permits.

\hypertarget{experimental-program}{%
\section{11. Experimental Program}\label{experimental-program}}

\hypertarget{rotary-phase-alignment-and-attention-score}{%
\subsection{11.1 Rotary-phase alignment and attention
score}\label{rotary-phase-alignment-and-attention-score}}

For controlled query--key pairs with magnitudes held fixed, uniformly
bounded wrapped pairwise phase displacement limits loss relative to full
alignment as stated in Lemma 1. The relevant quantities are

\[
\rho_{mnj}=|q_{m,j}||k_{n,j}|,\qquad
\delta_{mnj}=\alpha_{m,j}-\beta_{n,j}+(m-n)\theta_j. \tag{46}
\]

For a fully rotary head, the scaled score corresponding to Eq. (15) is

\[
\frac1{\sqrt{d_k}}\sum_j\rho_{mnj}\cos\delta_{mnj}. \tag{47}
\]

A numerical check of this identity or bound tests implementation and
stated assumptions. It is not, by itself, evidence that the proposed
hidden-state continuity measure predicts behavioral failure.

\hypertarget{long-range-continuity}{%
\subsection{11.2 Long-range continuity}\label{long-range-continuity}}

Task-relevant paired modal coordinates can be fitted or selected from
hidden-state trajectories. Hₜ = Γ(t,0) can then be tested as a predictor
of objective drift, contradiction, repetition, or unsupported
elaboration. The observable, anchor, basis, summaries, endpoint, and
treatment of undefined values must be fixed for each test.

\hypertarget{comparison-with-geometric-baselines}{%
\subsection{11.3 Comparison with geometric
baselines}\label{comparison-with-geometric-baselines}}

A proposed phase-coherence measure must be tested against alternatives
such as cosine similarity, subspace distance, centered kernel alignment,
linear probes, and activation-change metrics. A phase-coordinate
description earns predictive value only if it improves the specified
comparison; its notation alone supplies no advantage over geometry.

\hypertarget{positional-encoding-comparison}{%
\subsection{11.4 Positional-encoding
comparison}\label{positional-encoding-comparison}}

RoPE provides an explicit rotational decomposition in query--key
coordinates. Additive positional encoding does not introduce that same
operator by construction, although ordered hidden representations may
still admit spectral analysis. Matched comparisons should distinguish
properties of the positional operator from broader properties of ordered
trajectories.

\hypertarget{governance-effects-and-trajectory-diversity}{%
\subsection{11.5 Governance effects and trajectory
diversity}\label{governance-effects-and-trajectory-diversity}}

A governed execution layer can be tested for reduction of inadmissible
terminal transitions subject to a prespecified bound on loss of
trajectory diversity. Governed and ungoverned conditions should be
compared using defined boundary violations, task completion, diversity,
and continuity measures. A successful result would establish the
specified reduction and diversity bound under the tested conditions. It
would not establish general statistical or causal independence between
authority control and representational diversity.

\hypertarget{limits-of-the-framework}{%
\section{12. Limits of the Framework}\label{limits-of-the-framework}}

The framework does not establish that semantic structure is universally
harmonic, that Fourier modes are an optimal interpretive basis, or that
coherence is a universal predictor of hallucination. It does not equate
RoPE with physical wave propagation, quantum evolution, or a linear
time-invariant filter.

Its mathematical and methodological claims have distinct statuses:

\begin{enumerate}
\def\labelenumi{\arabic{enumi}.}
\tightlist
\item
  Ordered finite hidden-state sequences admit spectral decomposition.
\item
  RoPE implements position through rotations in paired coordinate
  planes.
\item
  Rotary scores decompose into magnitude-weighted cosine terms depending
  on content and relative position.
\item
  Lemma 1 bounds pre-softmax score loss relative to alignment under
  fixed magnitudes and bounded pairwise displacement.
\item
  Orthonormal paired coordinates define Γ under its stated weight and
  denominator conditions; useful semantic prediction requires separate
  evidence.
\item
  Representational continuity does not confer institutional
  authorization, and implemented authorization rules require separate
  evidence of legitimacy, evaluator adequacy, and enforcement.
\end{enumerate}

\hypertarget{future-research}{%
\section{13. Future Research}\label{future-research}}

Future work should distinguish mathematical properties, predictive
comparisons, causal interventions, and governance effects. None
substitutes for the others.

\hypertarget{basis-identifiability-and-invariant-phase-structure}{%
\subsection{13.1 Basis identifiability and invariant phase
structure}\label{basis-identifiability-and-invariant-phase-structure}}

Hidden-state phase requires a reproducible paired basis. Paired
principal or singular directions, task-trained orthogonal probes, and
the fixed feature-coordinate Fourier control defined in Section 6 can be
compared under a common evaluation protocol. Any data-dependent
construction must be fitted on training data; analytic controls must be
fixed without fitting. Operator-derived modes must be checked against
the orthonormality requirements of the functional or accompanied by a
separately justified construction.

Report coordinate-level phase and subspace-level invariants,
distinguishing common within-pair rotations from changes in subspace
selection and pairing. Repeated or nearly repeated singular values
require a declared treatment: sign conventions do not identify
directions within a degenerate singular subspace. Pairing across those
directions can affect Γ. Reproducibility and predictive comparisons must
account for this ambiguity.

\hypertarget{from-pairwise-score-bounds-to-attention-row-guarantees}{%
\subsection{13.2 From pairwise score bounds to attention-row
guarantees}\label{from-pairwise-score-bounds-to-attention-row-guarantees}}

Lemma 1 controls a single pre-softmax query--key score with fixed pair
magnitudes. A stronger theory would incorporate competing logits, masks,
magnitude perturbations, and margins. It would then ask whether a local
guarantee survives head aggregation, residual addition, normalization,
and multilayer propagation. Such results require additional assumptions;
they do not follow from the existing bound.

\hypertarget{causal-tests-of-continuity}{%
\subsection{13.3 Causal tests of
continuity}\label{causal-tests-of-continuity}}

Correlation between modal coherence and task completion would not
establish mechanism. Controlled interventions could rotate or attenuate
selected modal coordinates while preserving specified norms and
non-target subspaces, then measure objective retention, contradiction,
unsupported elaboration, repetition, and accuracy. Matched geometric
perturbations and random-subspace controls are needed to determine
whether effects are specific to the proposed construction.

\hypertarget{architecture-and-context-length-generalization}{%
\subsection{13.4 Architecture and context-length
generalization}\label{architecture-and-context-length-generalization}}

Matched-model studies should separate native rotary-coordinate structure
from spectral descriptions that apply to any ordered sequence. Within
RoPE systems, experiments can vary frequencies, scaling rules, sequence
length, and displacement range. The mathematical identity remains
conditional on the specified rotary operator; its behavioral usefulness
across these variations is a separate question.

\hypertarget{formal-separation-of-continuity-and-admissibility}{%
\subsection{13.5 Formal separation of continuity and
admissibility}\label{formal-separation-of-continuity-and-admissibility}}

Execution-boundary governance should be evaluated as an external
procedure over candidate transitions under an independently specified
contract. Governed and ungoverned conditions should be compared on
violations, false acceptance, false rejection, task completion, and
trajectory diversity. The hypothesis is that violations can be reduced
without unacceptable loss of representational variety or a uniform
coherence requirement. Coherence remains diagnostic, not evidence of
authority.

\hypertarget{falsification-criteria}{%
\subsection{13.6 Falsification criteria}\label{falsification-criteria}}

The framework should be rejected or narrowed if reproducible paired
bases cannot be identified, if phase measures fail to outperform
geometric baselines on held-out tasks, if targeted phase interventions
lack specific causal effects, or if the proposed governance separation
yields no measurable reduction in inadmissible transitions independent
of generic output suppression. These failure conditions are not
peripheral. They define the boundary between a useful spectral theory
and an attractive redescription of familiar geometry.

\hypertarget{empirical-follow-up}{%
\subsection{13.7 Empirical follow-up}\label{empirical-follow-up}}

A separate empirical report {[}9{]} evaluates selected continuity
features proposed within this program. The tested features did not
demonstrate the improvement required by the respective evaluation rules.
The methods, results, and limitations are reported there; those tests do
not evaluate the mathematical identity, the score-stability bound, or
CGM's enforcement effectiveness.

\hypertarget{conclusion}{%
\section{14. Conclusion}\label{conclusion}}

Transformer representations have ordered and rotational structure that
permits spectral description. Across context, hidden states form
sequences; within RoPE, relative position enters paired query--key
interactions as angular displacement. The resulting score is assembled
from magnitude-weighted cosine terms before masked softmax allocates
competitive attention.

The rotary decomposition and local bound establish properties of
pre-softmax compatibility under their assumptions. The proposed
hidden-state continuity measures require separate empirical support;
their mathematical definition alone does not establish predictive
usefulness.

Execution authorization remains a separate institutional requirement. A
coherent continuation can exceed delegated authority, while an
admissible continuation need not maximize representational similarity.
Governance equations specify a decision rule; effective implementation
requires accurate evaluation and control of the relevant execution
paths. Future work must test these mathematical, predictive, and
governance claims separately.

\hypertarget{acknowledgements}{%
\section{Acknowledgements}\label{acknowledgements}}

This work was developed independently at ENTRUST AI. The author has no
additional acknowledgements to declare.

\hypertarget{reproducibility-note}{%
\section{Reproducibility Note}\label{reproducibility-note}}

This paper presents mathematical constructions and a proposed
experimental program. It reports no experiment of its own. Mathematical
assumptions are stated in the relevant sections; the proposed
experiments should not be read as completed validation.

\hypertarget{conflict-of-interest-statement}{%
\section{Conflict of Interest
Statement}\label{conflict-of-interest-statement}}

The author is the founder and sole architect of ENTRUST AI, which
develops AI governance systems related to execution-boundary concepts
discussed in this paper. This affiliation constitutes a potential
commercial interest. CGM is discussed as an implementation example, not
as a product validated by the reported phase-prediction tests.

\hypertarget{references}{%
\section{References}\label{references}}

{[}1{]} A. Vaswani et al.,
\href{https://arxiv.org/abs/1706.03762}{``Attention Is All You Need,''}
\emph{Advances in Neural Information Processing Systems 30}, 2017,
pp.~5998--6008.
\href{https://doi.org/10.48550/arXiv.1706.03762}{\nolinkurl{doi:10.48550/arXiv.1706.03762}}
(arXiv version).

{[}2{]} J. Su, Y. Lu, S. Pan, A. Murtadha, B. Wen, and Y. Liu,
\href{https://arxiv.org/abs/2104.09864}{``RoFormer: Enhanced Transformer
with Rotary Position Embedding,''} arXiv:2104.09864, 2021, revised 2023.
\href{https://doi.org/10.48550/arXiv.2104.09864}{\nolinkurl{doi:10.48550/arXiv.2104.09864}}.

{[}3{]} I. Daubechies, ``The Wavelet Transform, Time-Frequency
Localization and Signal Analysis,'' \emph{IEEE Transactions on
Information Theory}, vol.~36, no. 5, pp.~961--1005, 1990.
\href{https://doi.org/10.1109/18.57199}{\nolinkurl{doi:10.1109/18.57199}}.

{[}4{]} S. Kornblith, M. Norouzi, H. Lee, and G. Hinton,
\href{https://proceedings.mlr.press/v97/kornblith19a.html}{``Similarity
of Neural Network Representations Revisited,''} \emph{Proceedings of the
36th International Conference on Machine Learning}, PMLR 97, 2019,
pp.~3519--3529.
\href{https://doi.org/10.48550/arXiv.1905.00414}{\nolinkurl{doi:10.48550/arXiv.1905.00414}}
(arXiv version).

{[}5{]} C. Olsson et al.,
\href{https://arxiv.org/abs/2209.11895}{``In-Context Learning and
Induction Heads,''} arXiv:2209.11895, 2022.
\href{https://doi.org/10.48550/arXiv.2209.11895}{\nolinkurl{doi:10.48550/arXiv.2209.11895}}.

{[}6{]} F. Barbero, A. Vitvitskyi, C. Perivolaropoulos, R. Pascanu, and
P. Veličković, \href{https://arxiv.org/abs/2410.06205}{``Round and Round
We Go! What Makes Rotary Positional Encodings Useful?''}
arXiv:2410.06205, 2024, revised 2025.

{[}7{]} Z. Gu, R. Chen, H. Zhang, H. Zhang, and Y. Hu,
\href{https://arxiv.org/abs/2505.13027}{``Deconstructing Positional
Information: From Attention Logits to Training Biases,''}
arXiv:2505.13027, rev. 2026.
\href{https://doi.org/10.48550/arXiv.2505.13027}{\nolinkurl{doi:10.48550/arXiv.2505.13027}}.

{[}8{]} F. Liu, \href{https://arxiv.org/abs/2602.10959}{``Rotary
Positional Embeddings as Phase Modulation: Theoretical Bounds on the
RoPE Base for Long-Context Transformers,''} arXiv:2602.10959, 2026.
\href{https://doi.org/10.48550/arXiv.2602.10959}{\nolinkurl{doi:10.48550/arXiv.2602.10959}}.

{[}9{]} A. Chachamovits, \href{https://arxiv.org/abs/2609.00335}{``Two
locked tests of phase-structure features for transition prediction,''}
arXiv:2609.00335, 2026. Accompanying revised manuscript:
\emph{Phase-structure features for contradiction and answer-change
prediction: A sealed comparison and retrospective development analysis},
September 2026.

\end{document}